\documentclass[10pt, a4paper]{article}

\usepackage[]{lrec-coling2024} 

\usepackage{xcolor}
\usepackage{hyperref}
 \definecolor{darkblue}{rgb}{0, 0, 0.5}
  \hypersetup{colorlinks=true, citecolor=darkblue, linkcolor=darkblue, urlcolor=darkblue}

\usepackage{xstring}

\usepackage{color}

\usepackage{wasysym}
\usepackage{subcaption}
\usepackage{soul}

\usepackage{amssymb}
\usepackage{xparse}
\NewDocumentCommand\smilingFaceWithHeartEyes{}{\includegraphics[scale=0.02]{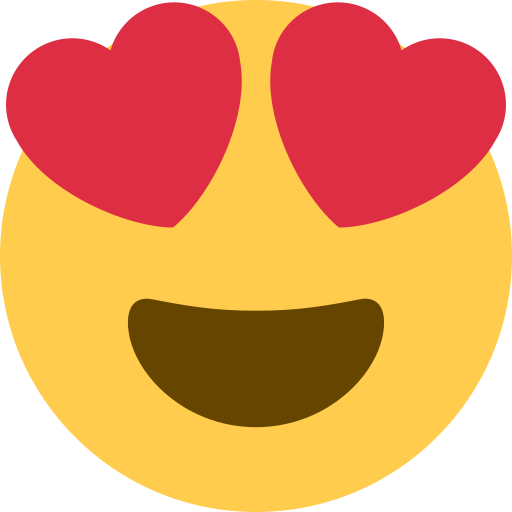}}
\NewDocumentCommand\loudylyCryingFace{}{\includegraphics[scale=0.02]{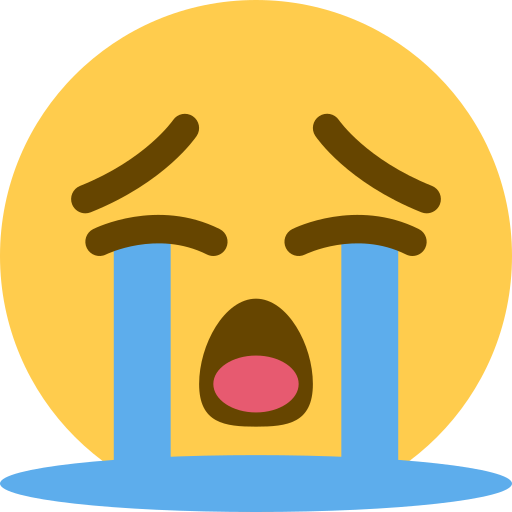}}
\NewDocumentCommand\smilingFaceWithSmilingEyes{}{\includegraphics[scale=0.02]{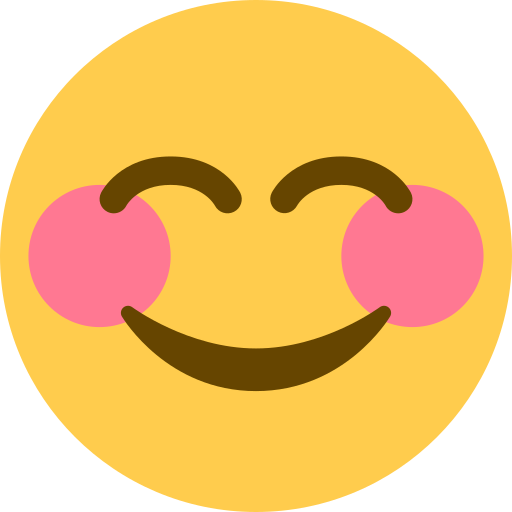}}
\NewDocumentCommand\upsideDwonFace{}{\includegraphics[scale=0.02]{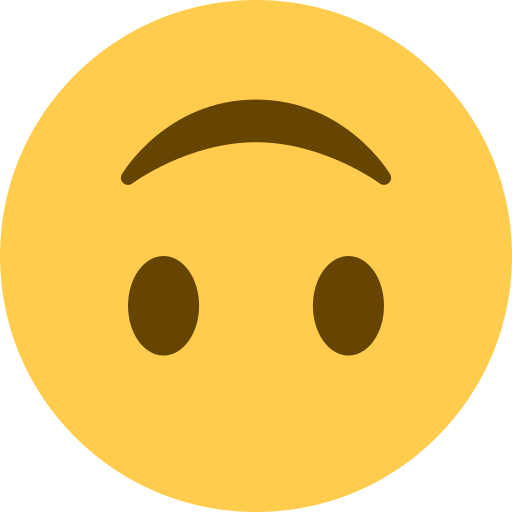}}
\NewDocumentCommand\faceWithTearsOfJoy{}{\includegraphics[scale=0.02]{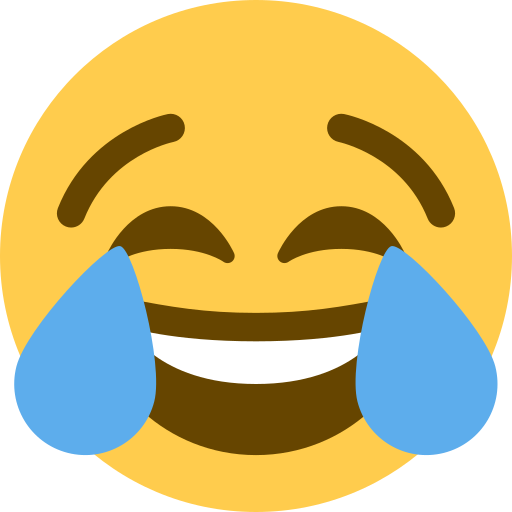}}
\NewDocumentCommand\sleepyFace{}{\includegraphics[scale=0.02]{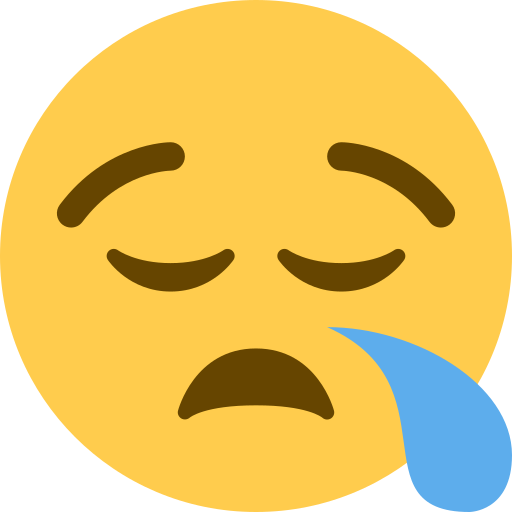}}
\NewDocumentCommand\smilingFace{}{\includegraphics[scale=0.02]{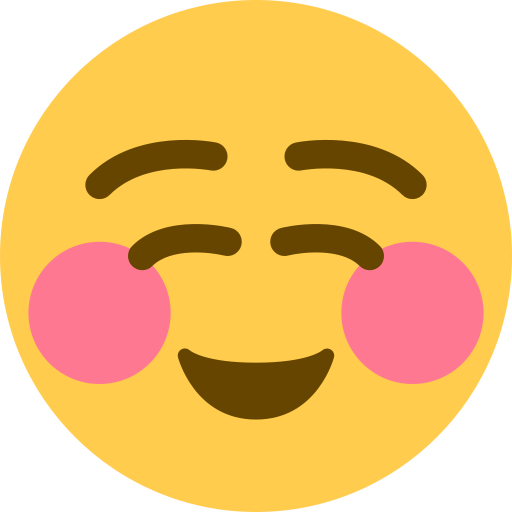}}
\NewDocumentCommand\pleedingFace{}{\includegraphics[scale=0.02]{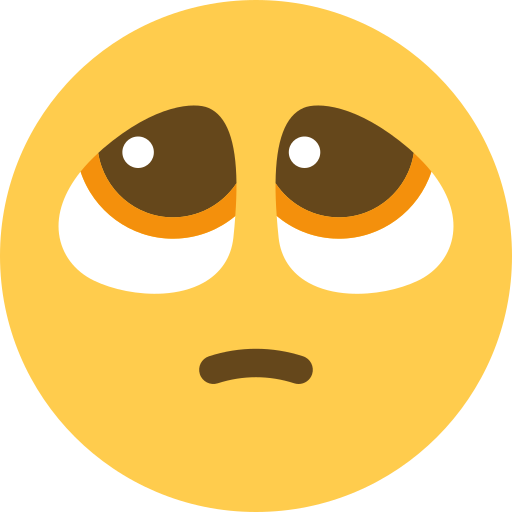}}

\NewDocumentCommand\beamingFaceWithSmilingEyes{}{\includegraphics[scale=0.02]{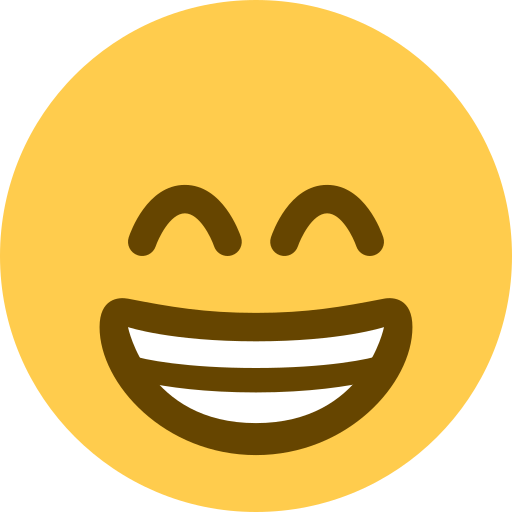}}
\NewDocumentCommand\grinningFaceWithSweat{}{\includegraphics[scale=0.02]{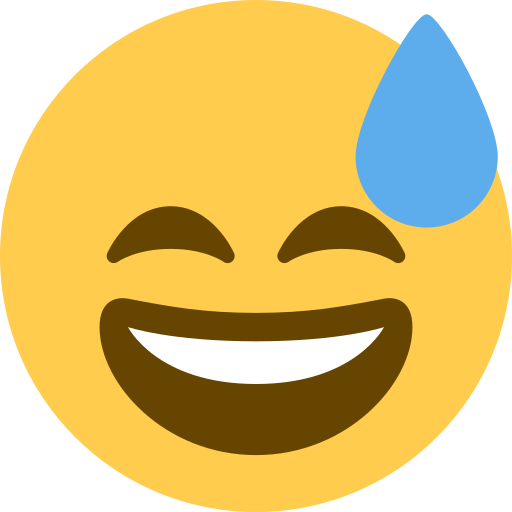}}
\NewDocumentCommand\personTippingHand{}{\includegraphics[scale=0.02]{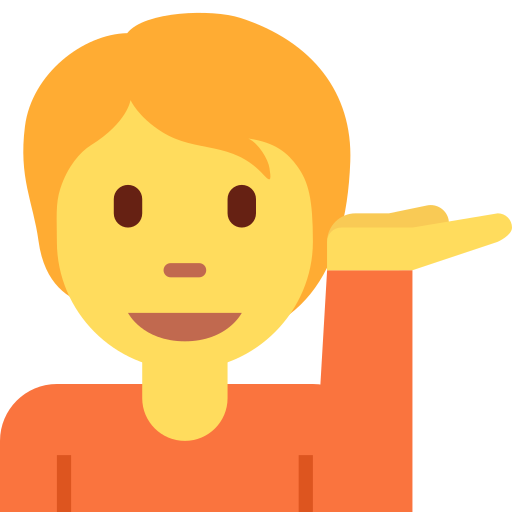}}

\NewDocumentCommand\flushedFace{}{\includegraphics[scale=0.02]{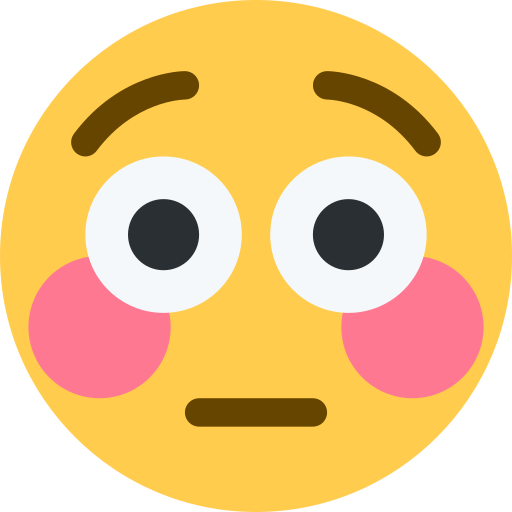}}

\NewDocumentCommand\okHand{}{\includegraphics[scale=0.02]{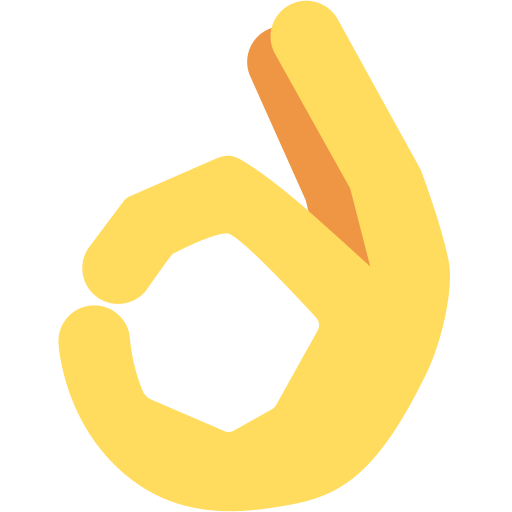}}
\NewDocumentCommand\redHeart{}{\includegraphics[scale=0.02]{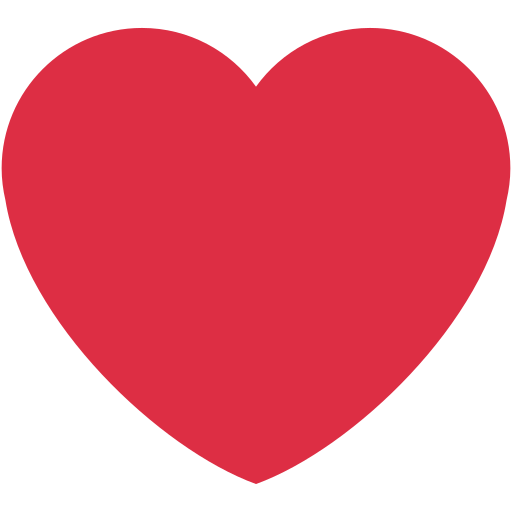}}

\NewDocumentCommand\blueHeart{}{\includegraphics[scale=0.02]{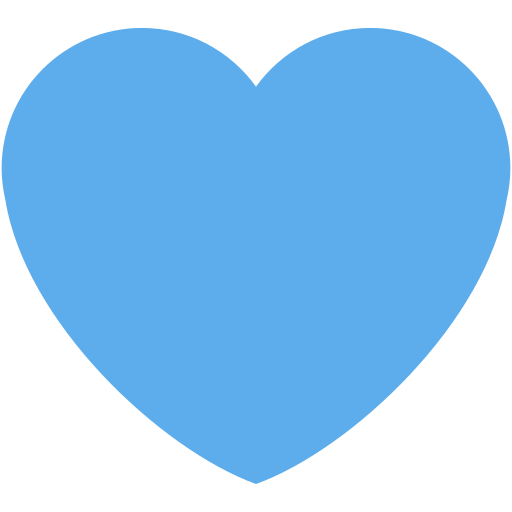}}
\NewDocumentCommand\seeNoEvilMonkey{}{\includegraphics[scale=0.02]{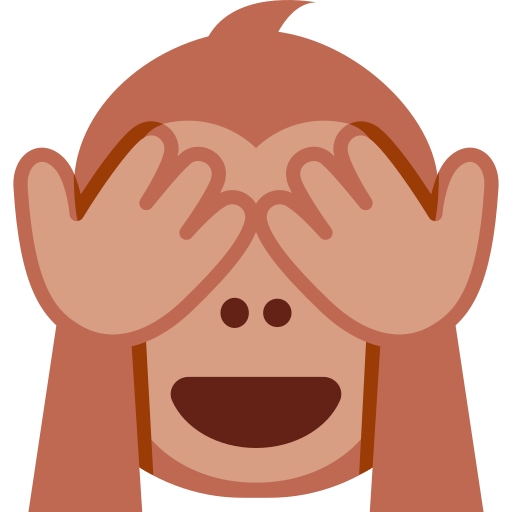}}

\NewDocumentCommand\purpleHeart{}{\includegraphics[scale=0.04]{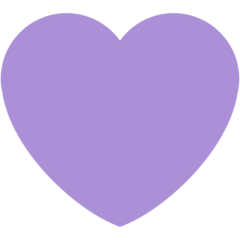}}

\title{Who is bragging more online? A large scale analysis of bragging in social media}

\name{Mali Jin\textsuperscript{$\diamondsuit$}, Daniel Preo\c{t}iuc-Pietro\textsuperscript{$\spadesuit$}, A. Seza Doğruöz\textsuperscript{$\clubsuit$}, Nikolaos Aletras\textsuperscript{$\diamondsuit$}} 

\address{\textsuperscript{$\diamondsuit$}University of Sheffield, \textsuperscript{$\spadesuit$}Bloomberg, \textsuperscript{$\clubsuit$} LT3, IDLab, Universiteit Gent \\
\texttt{\{m.jin, n.aletras\}@sheffield.ac.uk} \\
\texttt{dpreotiucpie@bloomberg.net} \\
\texttt{as.dogruoz@ugent.be} \\}

\abstract{
Bragging is the act of uttering statements that are likely to be positively viewed by others and it is extensively employed in human communication with the aim to build a positive self-image of oneself. Social media is a natural platform for users to employ bragging in order to gain admiration, respect, attention and followers from their audiences. Yet, little is known about the scale of bragging online and its characteristics. This paper employs computational sociolinguistics methods to conduct the first large scale study of bragging behavior on Twitter (U.S.) by focusing on its overall prevalence, temporal dynamics and impact of demographic factors. Our study shows that the prevalence of bragging decreases over time within the same population of users. In addition, younger, more educated and popular users in the U.S. are more likely to brag. Finally, we conduct an extensive linguistics analysis to unveil specific bragging themes associated with different user traits. 
 \\ \newline  }

\begin{document}

\maketitleabstract

\section{Introduction}

Bragging (self-praise) is a speech act that involves disclosing positive content about oneself or their close network~\cite{dayter2014self,dayter2018self} such as achievements, possessions or feelings~\cite{scopelliti2015you,matley2018not}. It falls in the category of self-presentation strategies with the goal of building and establishing a positive social image of one self~\cite{goffman1978presentation,jones1982toward,jones1990interpersonal,bak2014self}. This desire for attaining a positive self-image is a key driver of human behavior~\cite{baumeister1982self,leary1990impression,sedikides1993assessment,tetlock2002social}, as it can bring benefits like admiration, respect, attention by other people and personal rewards~\cite{gilmore1989effects,hogan1982socioanalytic,schlenker1980impression}. Bragging as a behavior was found to be related to user traits including personality and it can differ based on communication strategies favored by the participants to the bragging act~\cite{miller1992should,van2017praising,sezer2018humblebragging}.

Social media presents a unique environment to study bragging (and self-presentation strategies more generally) as users' craft and control the information they expose to their audience. Further, due to the nature of online interactions, this information can represent a significant part of what most of the audience knows about a person. Thus, users on social media platforms pay great attention to building their online social image~\cite{chou2012they,michikyan2015can,halpern2017online} and studies have shown that online self-presentation is predominantly positive~\cite{matley2018not,chou2012they,lee2014puts}. Moreover, social media platforms include functionality that rewards and promotes positive statements such as likes or positive reactions to users’ posts~\cite{reinecke2014authenticity} which often are used to quantify impact on the platform~\cite{lampos-etal-2014-predicting,mu2024predicting}. Traditionally, bragging is considered a high risk act~\cite{van2017praising,brown1987politeness,holtgraves1990language} and can lead to the opposite effect than intended, such as dislike or decreased perceived competence~\cite{matley2018not,jones1982toward,sezer2018humblebragging}. However, bragging is acceptable and even desired in certain online contexts~\cite{dayter2018self}. According to ~\citet{dayter2018self,matley2020isn,ren2020self} self-promotion is pervasive on social media. Bragging in particular is more frequent on social media than face-to-face interactions~\cite{ren2020self}. Table \ref{t:bragging} shows examples of bragging and non-bragging tweets.

\renewcommand{\arraystretch}{1.2}
\begin{table}
\centering
\small
\begin{tabular}{|l|p{0.3\textwidth}|}
\hline
\bf Type & \bf Text \\ \hline
Bragging &  \textit{Just impressed myself with how much French I think I understood! One semester at KC FTW!} \\
\hline
Non-bragging & \textit{Glad to hear that! Well done Jim!} \\
\hline
  \end{tabular}
\caption{\label{t:bragging} Example of bragging and non-bragging tweets.}
\end{table}

To date, bragging behavior was quantitatively studied in specific contexts through manual analyses of small data sets of hundreds of posts~\cite{dayter2014self, dayter2018self,matley2018not,tobback2019telling,rudiger2020manbragging}.
This paper aims to provide the first large scale quantitative study on the prevalence of online bragging using a longitudinal data set that includes over 1 million English tweets posted by a group of 2,685 Twitter users in the U.S. over ten years. We identify bragging through Natural Language Processing (NLP) methods, including using a state-of-the-art neural model~\cite{jin2022automatic} for identifying bragging in social media posts. Our main findings are as follows:
\begin{itemize}
    \item Bragging steadily decreases over time in the past 10 years within the same group of users.
    \item Bragging is more prevalent among users who are female, generally younger, more educated, have a higher income and are more popular on Twitter.
    \item Male users and users with higher income brag more about leisure activities. Female users and users with lower education focus more on themselves when bragging. Bragging by older users and users who have higher education are more likely to involve others. Emojis are more frequently used by female and younger users while bragging. 
\end{itemize}

\section{Related Work}

\paragraph{Bragging as a Social Behavior}
Bragging enhances one's positive social image and can trigger high-quality interactions with other users \citep{miller1992should}. Therefore, it is acceptable and even desirable on social media to some extent \citep{dayter2018self}. However, bragging is also widely considered as a face-threatening act for speakers according to the modesty maxim \citep{leech2016principles} and politeness theory as it threatens both positive and negative face \citep{brown1987politeness}. As a result, it may damage user likability if it is negatively perceived by the audience \citep{matley2018not}. Thus, social psychology and linguistic studies have mostly focused on investigating the impact of self-promotion for identifying its pragmatic strategies (e.g., shifting focus to others who are closely related to them). These strategies help the speakers/users to mitigate the social risk and negative effects caused by bragging \citep{scopelliti2015you,tobback2019telling,matley2020isn}.

\paragraph{Analysis of Bragging}
The use of pragmatic strategies for bragging has been analyzed qualitatively by linguists and psychologists across languages such as Mandarin \citep{wu2011conversation}, English \citep{speer2012interactional,dayter2021dealing}, Spanish \citep{maiz2021blowing}, or Russian \citep{dayter2021dealing}. In social media, \citet{dayter2014self} identified a series of overlapping strategies in a small ballet community on Twitter. \citet{matley2018not} examined the pragmatic function of hashtags (e.g., \#brag, \#humblebrag) used by Instagram users and \citet{tobback2019telling} studied the impact of bragging about professional skills on LinkedIn. Also, the frequency of different bragging strategies used by regular people \citep{ren2020self} and celebrities \citep{guo2020managing} has been examined in Weibo (a Chinese social media platform). Furthermore, the emotional influence of bragging from the audience perspective was studied by \citet{scopelliti2015you}. More recently, \citet{jin2022automatic} introduced the task of automatically identifying bragging and classifying its types on Twitter using computational methods.

\paragraph{Language Use and User Traits}
Previous work in computational sociolinguistics showed that user traits (e.g., age, gender and personality) and popularity correlate with language use and online behavior \citep{preotiuc-pietro-etal-2015-analysis,nguyen2016computational,preotiuc-pietro-etal-2017-beyond,preotiuc-pietro-ungar-2018-user,villegas2023multimodal}. 
\citet{schwartz2013personality} found significant variations in language on Facebook with age, gender and personality while \citet{preoctiuc2015studying} examined the correlations between user income and a variety of factors such as profile features, demographic and psychological traits.


Studies on differences in self-disclosure have focused on gender through different platforms \citep{valkenburg2011gender,sheldon2013examining,altenburger2017there,farinosi2018can,wang-etal-2021-self}, where most of them have suggested that women disclose more about themselves than men do. Also, racial background was investigated in self-disclosure behavior~\citep{chen1995differences,rui2013strategic}. \citet{bak2012self} investigated the relationship between self-disclosure and user relationship strength (e.g., strong and weak) on Twitter, while \citet{moon2016role} examined the relationship between narcissism (i.e., a personality trait reflecting a grandiose and inflated self-concept) and self-promoting on Instagram. 
To the best of our knowledge, no previous study has attempted to study the relationship between bragging and social factors.

\section{Measuring Bragging Prevalence}

\subsection{Data for Model Training}

We use a publicly available data set developed by~\citet{jin2022automatic} to train a bragging classifier and it consists of 6,696 English tweets and each tweet is manually annotated as bragging or non-bragging. In total, there are 544 bragging tweets and 2,838 non-bragging tweets in the training set; 237 bragging tweets and 3,077 non-bragging tweets in the development and test set (see Table \ref{t:data_statistics}).\footnote{This data set is also manually annotated according to six bragging types (i.e., \textit{Achievement}, \textit{Action}, \textit{Feeling}, \textit{Trait}, \textit{Possession}, \textit{Affiliation}). We only use the binary label because of the small number of examples in each bragging type which results in low performance on type prediction. This can introduce noise in our analysis. } 

\renewcommand{\arraystretch}{1.2}
\begin{table}
\centering
\resizebox{0.48\textwidth}{!}{
\begin{tabular}{|l|c|c|c|}
\hline
\bf Label & \bf Training set & \bf Dev/Test set &\bf All \\ \hline
Bragging & 544 (16.09\%) & 237 (7.15\%) & 781 (11.66\%)  \\
Non-bragging & 2,838 (83.91\%) & 3,077 (92.85\%) & 5,915 (88.34\%) \\
\hline
Total & 3,382 & 3,314 & 6,696 \\
\hline
\end{tabular}}
\caption{\label{t:data_statistics} Statistics of the bragging data set.}
\end{table}

\subsection{Predictive Model}

We re-implement the best performing predictive model proposed by~\citep{jin2022automatic} on identifying whether a tweet contains bragging or not (BERTweet-LIWC).\footnote{We achieve similar performance on the bragging test set (72.51 macro F1).} 
The BERTweet-LIWC model generates word embeddings using BERTweet~\cite{nguyen2020bertweet} and combines it with Linguistic Inquiry and Word Count (LIWC; ~\cite{pennebaker2001linguistic}) features. The feature combination is performed through a fusion mechanism, which was originally introduced by~\citet{rahman2020integrating}, to control the influence of word representations and linguistic features. The joint representation of text and LIWC features is sent to the BERTweet encoder, followed by an output binary classification layer. 

\subsection{Bragging Prevalence Metrics}

Our analyses rely on computing a user-level bragging prevalence score. We expect that the base rate of bragging will change over time. However, we aim to compare groups of users based on their traits and we want to ensure the bragging score is adjusted to the base rate in bragging over a certain time period. Thus, we normalize the score over several smaller time chunks.

\paragraph*{Overall Bragging Prevalence}

First, we obtain the time series of total bragging tweets and total tweets posted by all users over each month, denoted as $P = \{p_1, ..., p_n\}$ and $Q = \{q_1, ..., q_n\}$ respectively where $n$ is the number of months in the data set. We compute the distribution of average bragging percentage across all users for each month $A = \{a_1, ..., a_n\}$ such that:
\begin{eqnarray}
\footnotesize
A = \frac{P}{Q}
\label{equ:allUsers}
\end{eqnarray}

\begin{figure}[!t]
\center
\includegraphics[scale=0.5]{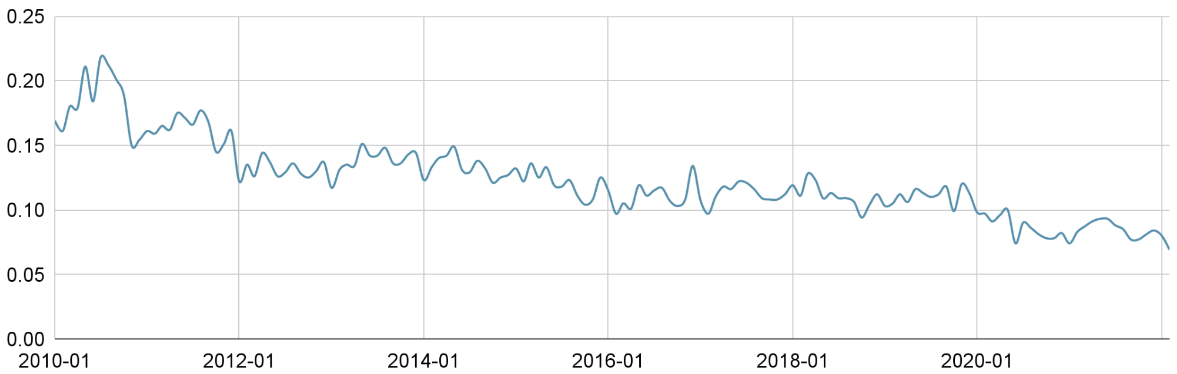}
\caption{Bragging percentage by year and month.}             
\label{fig:bragging_year_and_month} 
\end{figure}

\paragraph*{User Bragging Prevalence} To compute a bragging score of an individual user $u$, we first obtain the distributions of bragging tweets and total tweets over each  month for $u$, which are denoted as $B_u = \{{b_u}_1, ..., {b_u}_n\}$ and $T_u = \{{t_u}_1, ..., {t_u}_n\}$ respectively. We obtain a time-normalized bragging distribution $D_u = \{d_{u1}, ..., d_{un}\}$ for each user over months by dividing each data point from the user distribution by the overall bragging prevalence in each time window:
\begin{eqnarray}
\footnotesize
D_u = \frac{B_u}{T_u*A}
\label{equ:individualUsers}
\end{eqnarray}
Finally, we average the normalized bragging distribution $D_u$ for each user $l$:
\begin{eqnarray}
\footnotesize
l = \frac{\sum_{i=1}^n {d_u}_i}{n}
\label{equ:individuaAverage}
\end{eqnarray}
The normalized bragging percent practically compares the bragging tendency of a single user to that of the population average in the same time range. We normalize with time to account for possible temporal shifts in bragging prevalence over time (see Figure~\ref{fig:bragging_year_and_month}).

\section{Analysis Data}

To analyze the social factors of bragging, we need a large set of Twitter users associated with socio-demographic characteristics. We combine three data sets that contain such information in the following papers as provided by the original authors on request: the first data set (developed by~\citealt{preotiuc2016studying}) contains 863 users, the second data set (developed by~\citealt{guntuku2017studying}) contains 4,568 users and the third data set (developed by~\citealt{jaidka2020estimating}) contains 938 users. All user information is obtained through self-reports on questionnaires. All users are mapped to their self-reported gender and age, while the users in the second data set also self-report their education degree and annual income range. In total, our combined data set contains 6,369 users, all from the U.S.. For details about user sampling, please refer to the original papers. Subsequently, we collect all historical tweets from these users resulting in more than 9.7 million tweets. We stop data collection at the end of 2021.

\subsection{Filtering}

For computing user bragging behavior, we need to only study content that is original and authored by the user. For this, we employ the following filtering steps:
\begin{itemize}
    \item We filter out non-English content using the language code provided by Twitter, as our classifier is only trained to classify English content.
    \item We exclude replies and retweets through the parameters exposed through the Twitter API.
    \item We remove duplicate tweets from the same user by using the first five content tokens (excluding numbers, usernames and hashtags), because duplicate tweets are unlikely to be original content created by users. 
    \item All tweets that are automatically generated from third parties are filtered out using source labels that indicate original authorship such as ``Twitter Web Client`` or ``Twitter for Android". We found that tweets with source labels such as ``The Sims 4 Game" and ``Paradise Island 2" are likely to be generated automatically and would negatively impact our analysis (e.g., \textit{``I played the Sandy Caps mini game in Paradise Island 2, and my score was: 68 \#ParadiseIsland2 \#GameInsight''}).
    \item Finally, we remove all users that have posted fewer than 20 tweets, as computing an average bragging ratio across a very small number of tweets results in unreliable estimates.
\end{itemize}

In total, our analysis data contains 2,685 users with 1,031,276 tweets, with each user having 384 posts on average. 

\subsection{Text Processing}
We pre-process the collected tweets by lower-casing and tokenizing using TweetTokenizer from NLTK Toolkit~\cite{bird2009natural}. We also replace all URLs and username mentions with a single word token <URL>  and <USER> respectively. 

\subsection{Computing Bragging Ratio}

We use the bragging predictive model to identify the category (bragging or non-bragging) of all 1,031,276 individual tweets in our analysis data set. Then we compute the normalized bragging ratio for each user using Equation~\ref{equ:individualUsers} and Equation~\ref{equ:individuaAverage}.

To investigate the performance of the model on the analysis data, we manually evaluated a batch of 100 tweets across different users and years and found that the model achieved 78.55 macro F1, which was a higher absolute performance number on the test data set than the one reported in~\citet{jin2022automatic} (72.51 macro F1). Also, we notice that it is easier for the model to misclassify non-bragging tweets as bragging in the analysis data set due to positive sentiment or gratitude expressions in the text (e.g., \textit{``Just installed box.net iOS app and signed up for a FREE 50gb lifetime cloud space account . Thx <USER>! \#gaetc \#gaetcsmackdown \#freeapp''}); while bragging tweets are misclassified as non-bragging are mostly due to indirect expressions (e.g., \textit{``\#MyBestFriendsKnow that I am that one friend <URL>''}). This is in line with the error analysis in \citet{jin2022automatic}. In general, mispredicted tweets are evenly distributed over the entire data coverage period except for the years 2012 and 2018. Among these error cases, 40\% of tweets are posted by males and 60\% of these are posted by females; while 55\% of them are generated by users born before 1988 and 45\% of them are generated by those born after 1988.

\subsection{User Demographic Traits}
\label{ssec:demographic}

To examine the relationship between bragging behavior online and the following user demographic traits, we use the following attributes.

\begin{itemize}
    \item \textit{Gender}. There are 859 self-identified males (33.81\%) and 1,678 self-identified females (66.19\%). Self-identified non-binary users represented a very small number of the total participants, hence were removed from the analysis.
    \item \textit{Age}. Reported as the year of birth. The age reported throughout the paper refers to user age as of the end of 2021.
    \item \textit{Education}. It contains six categories: (1) users who have not completed high school; (2) high school or equivalent; (3) associate's degree or equivalent; (4) bachelor's degree or equivalent; (5) master's degree or equivalent; and (6) doctoral degree or equivalent.
    \item \textit{Income}. The annual yearly income of a user in U.S. dollars is divided into eight categories: (1) below 10K; (2) 10-25K; (3) 25-40K; (4) 40-60K (5) 60-75K; (6) 75-100K; (7) 100-200K; and (8) above 200K.
\end{itemize}

\subsection{User Popularity Traits}

Popularity reflects other people's interest in users' accounts or posts. We quantify this using the following metrics: (1) the number of followers (the higher, the more popular); (2) the ratio between friends or users being followed and followers (the lower, the more popular) and (3) the number of times a user was listed (the higher, the more popular).
Note that, for computing the correlation between popularity metrics and bragging, we log scale the values such that their distribution are closer to a Gaussian distribution. We collect the popularity user information at the end of 2021, consistent with the cut-off date for the data collection.

\begin{figure*}[t!]
     \centering
     \begin{subfigure}[b]{0.28\textwidth}
         \centering
         \includegraphics[width=\textwidth]{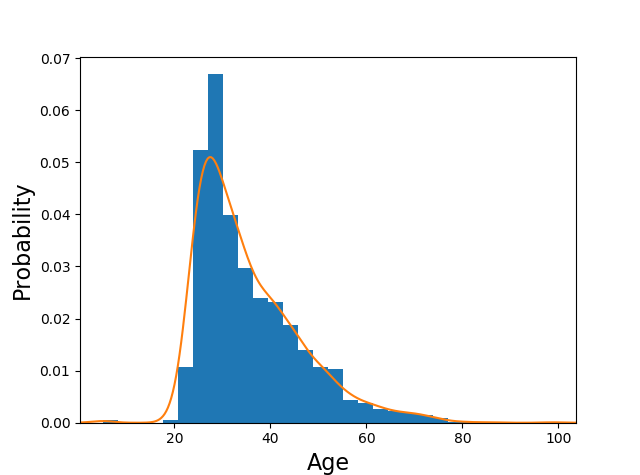}
         \caption{Age.}
         \label{subfig:histogram_age}
     \end{subfigure}
     \hfill  
     \begin{subfigure}[b]{0.28\textwidth}
         \centering
         \includegraphics[width=\textwidth]{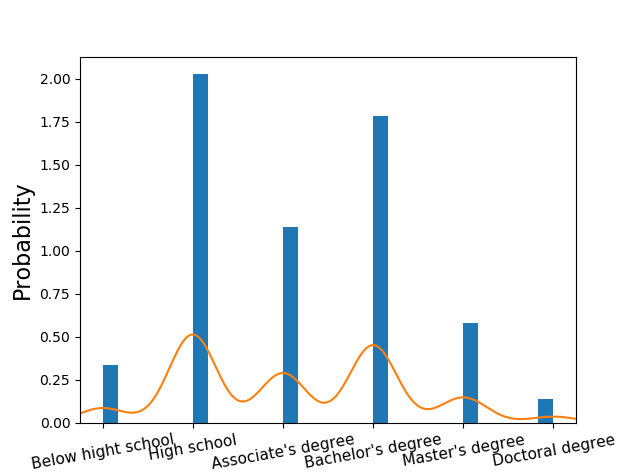}
         \caption{Education.}
         \label{subfig:histogram_education}
     \end{subfigure}
     \hfill
     \begin{subfigure}[b]{0.28\textwidth}
         \centering
         \includegraphics[width=\textwidth]{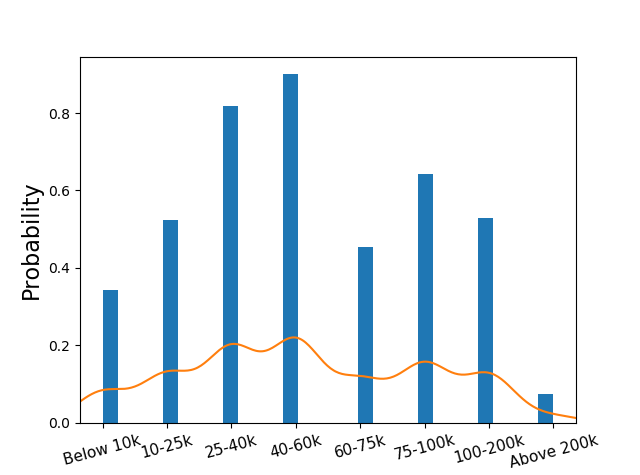}
         \caption{Income.}
         \label{subfig:histogram_income}
     \end{subfigure}
     \hfill
     \begin{subfigure}[b]{0.28\textwidth}
         \centering
         \includegraphics[width=\textwidth]{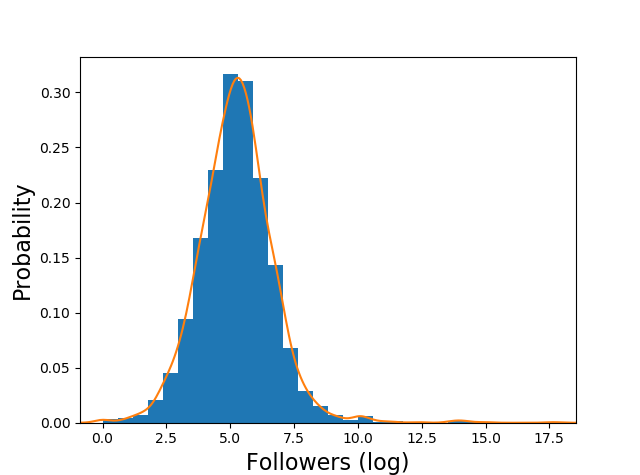}
         \caption{Log-scaled no. followers.}
         \label{subfig:histogram_follower}
     \end{subfigure}
     \hfill
     \begin{subfigure}[b]{0.28\textwidth}
         \centering
         \includegraphics[width=\textwidth]{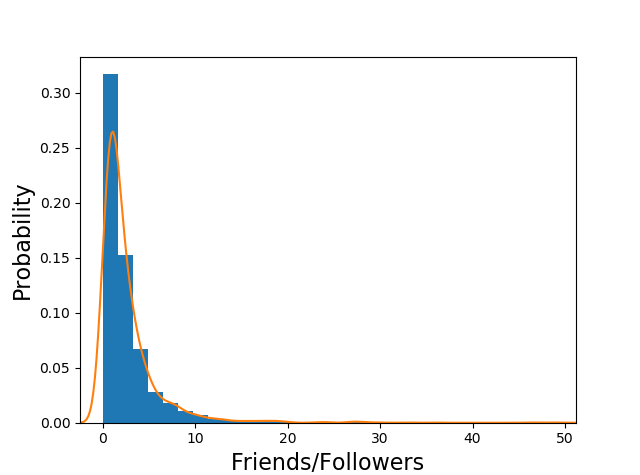}
         \caption{Friends/followers.}
         \label{subfig:histogram_friend_follower}
     \end{subfigure}
     \hfill
     \begin{subfigure}[b]{0.28\textwidth}
         \centering
         \includegraphics[width=\textwidth]{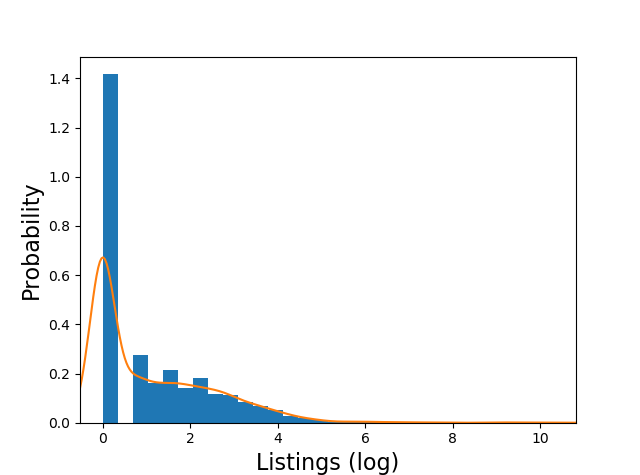}
         \caption{Log-scaled no. listings.}
         \label{subfig:histogram_listing}
     \end{subfigure}  
        \caption{Histograms of user socio-demographic traits and popularity.}
        \label{fig:histograms}
\end{figure*}

Table~\ref{t:feature_statistics} shows statistics of the demographic and popularity traits in our data. The trait distributions in the data set are presented in Figure~\ref{fig:histograms}.

\renewcommand{\arraystretch}{1.2}
\begin{table}
\centering
\resizebox{0.48\textwidth}{!}{
\begin{tabular}{|l|c|c|}
\hline
\bf Trait & \bf Mean & \bf Median \\ \hline
\multicolumn{3}{|l|}{Demographic Traits} \\
\hline
Age & 35.99 & 33 \\
\hline
\multicolumn{3}{|l|}{User Impact} \\
\hline
No. Followers & 22,951.06 & 186 \\
No. Friends/No. Followers & 2.62 & 1.58\\
No. Listings & 47.72 & 2.0\\
\hline
\end{tabular}}
\caption{\label{t:feature_statistics} Statistics of user socio-demographic traits and popularity.}
\end{table}


\section{Results and Discussion}

\subsection{Bragging Prevalence over Time}

We first look at the frequency of bragging on Twitter through our data set and its dynamics over time. The mean bragging ratio is 0.1114 and the median bragging ratio is 0.0968 (0.0020 and 0.0013 after normalization) for all users. This shows that bragging is a common occurrence in original tweets from U.S. users, with users bragging on average in 1 out of 9 of their tweets.
Next, we study the temporal dynamics of bragging. Figure~\ref{fig:bragging_year_and_month} shows the bragging percentage over time across the ten years of the data set. This shows that overall the bragging percentage is slowly decreasing with time. The decline in bragging behavior on Twitter during the study period may be attributed to various factors such as a shift in focus to other social media platforms. Thus, it would be challenging to ascertain the exact reasons based solely on our Twitter data. Additionally, this finding highlights the need for temporal normalization of the bragging ratio, as users who published more recently could be skewed to having a lower overall bragging ratio.

\subsection{When to Brag and How}

Next, we investigate the time (e.g., day of the week, time of the day) of bragging and the linguistic patterns around these times. Overall, we do not find any significant correlations between bragging rate and either the time of day, when split into ranges, or the day of the week (e.g., weekday vs. weekend $\rho=0.02$).

However, even if the bragging rate is similar, we also explore if the expressions of bragging across days of the week and times of the day vary using a linguistic analysis. For this, we use a unigram (i.e., token) feature analysis to identify words and themes associated with bragging by computing the correlations between the distribution of each unigram across posts and the label of the post (i.e., bragging or not bragging). Then, we use univariate Pearson correlation to rank the unigrams similar to~\cite{schwartz2013personality}.

\paragraph{Day of the Week}
Table~\ref{t:feature_weekday_weekend} presents the top-20 unigram features correlated with bragging in weekdays vs. weekends. For this analysis, we normalize the creation time of each post to the local time by using the timezone difference which was inferred from the zip code that users have provided.

\renewcommand{\arraystretch}{1.2}
\begin{table}[!t]
\center
\begin{tabular}{|l|c|l|c|}
\hline
\multicolumn{4}{|c|}{\bf Weekday vs. Weekend}  \\
\hline
Unigram & r & Unigram & r  \\ 
\hline
class & .050 & sunday & .034 \\
professor & .028 & night & .029\\
semester & .028 & friday & .028\\
job & .026 & church & .028\\
campus & .024 & <user> & .026\\
classes & .023 & \#livepd & .025\\
\#tbt & .023 & weekend & .025\\
thursday & .021 & we & .024\\
interview & .021 & state & .024\\
exam & .021 & watching & .020\\
office & .020 & bar & .020\\
bio & .020 & game & .020\\
monday & .020 & won & .019\\
killed & .019 & football & .019\\
grow & .019 & drinking & .019\\
vote & .019 & rewards & .018\\
internship & .018 & afternoon & .018\\
ago & .018 & party & .018\\
teeth & .017 & racing & .018 \\
gym & .017 & jam & .017\\
\hline
\end{tabular}
\caption{Unigram feature correlations with bragging between weekday and weekend, sorted by Pearson correlation (r). All correlations are significant at $p < .001$, two-tailed t-test.}
\label{t:feature_weekday_weekend}
\end{table}

To demonstrate the face validity of the analysis, we first observe that bragging statements from both weekdays and weekends involve words related to time (e.g., \textit{thursday}, \textit{sunday}, \textit{night}, \textit{weekend}, \textit{afternoon}). Secondly, we observe that users mostly brag about their school life or work on weekdays (e.g., \textit{class}, \textit{professor}, \textit{interview}, \textit{office}, \textit{internship}). For example, a user mentions his/her professor when bragging.
\begin{quote}
    \footnotesize
    T1: \textit{``My professor for accounting saw what I carry all day and said I must have great upper body strength lol''}.
\end{quote}
Another popular bragging topic on weekdays is about going to the \textit{gym}. Bragging on weekends usually focuses on certain entertainment, recreation and worship activities (e.g., \textit{church}, \textit{watching}, \textit{bar}, \textit{football}, \textit{drinking}, \textit{party}) or church attendance (\textit{church}) (see example T2). 
\begin{quote}
    \footnotesize
    T2: \textit{``Yay, I'm picking up my cute new glasses tomorrow. Now I can rock them at Mom's party tomorrow''}.
\end{quote} 
In addition, this could also involve activities that are done as part of a group, as exemplified by the first person plural pronoun \textit{we}.

\paragraph*{Time of Day} Table~\ref{t:feature_time_in_day} shows the unigrams associated with bragging at different times in a day. Similar to the day of the week analysis, our findings demonstrate face validity since the top correlated tokens are related to the time of bragging (e.g., \textit{morning}, \textit{yesterday}, \textit{tonight}, \textit{2:30}).

\renewcommand{\arraystretch}{1.3}
\begin{table*}[!t]
\small
\center
\resizebox{\textwidth}{!}{
\begin{tabular}{|l|c|l|c|l|c|l|c|l|c|l|c|l|c|l|c|}
\hline
\multicolumn{2}{|c|}{\bf 06:00-09:00} & \multicolumn{2}{|c|}{\bf 09:00-12:00} & \multicolumn{2}{|c|}{\bf 12:00-15:00} & \multicolumn{2}{|c|}{\bf 15:00-18:00} & \multicolumn{2}{|c|}{\bf 18:00-21:00} & \multicolumn{2}{|c|}{\bf 21:00-00:00} & \multicolumn{2}{|c|}{\bf 00:00-03:00} & \multicolumn{2}{|c|}{\bf 03:00-6:00}\\
\hline
Unigram & r & Unigram & r & Unigram & r & Unigram & r & Unigram & r & Unigram & r & Unigram & r & Unigram & r\\ 
\hline
morning & .131 & morning & .042 & lunch & .053 & dinner & .040 & dinner & .042 & tomorrow & .049 & i & .055 & sleep & .069\\
last & .055 & coffee & .040 & just & .029 & $<$url$>$ & .025 & \#piano & .031 & tonight & .047 & night & .044 & awake & .069\\
breakfast & .055 & class & .037 & afternoon & .023 & store & .022 & \#music & .026 & bed & .040 & sleep & .043 & morning & .069\\
today & .051 & last & .031 & shopping & .020 & pizza & .021 & tonight & .024 & i & .038 & midnight & .041 & early & .058\\
day & .048 & today & .029 & basically & .020 & came & .021 & beer & .022 & night & .029 & bed & .032 & up & .057\\
up & .046 & lunch & .028 & outside & .020 & bags & .020 & tomorrow & .020 & \#livepd & .026 & \sleepyFace{}{} & .032 & 4am & .051\\
sleep & .044 & yesterday &.028 & shop & .020 & fitness & .018 & \#musicmonday & .020 & midnight & .026 & 2:30 & .031 & 5:30 & .047\\
coffee & .044 & woke & .028 & break & .020 & mail & .018 & pizza & .021 & win & .025 & drunk & .031 & 6:30 & .042\\
school & .044 & breakfast & .027 & cleaning & .019 & published & .018 & \#classicalmusic & .021 & life & .025 & tonight & .030 & covet & .041\\
yesterday & .043 & classes & .025 & food & .019 & rain & .017 & wine & .021 & absolutely & .023 & friends & .030 & singles & .039\\
\hline
\end{tabular}}
\caption{Unigram feature correlations with bragging between different time periods in a day, sorted by Pearson correlation (r). All correlations are significant at $p < .001$, two-tailed t-test.}
\label{t:feature_time_in_day}
\end{table*}

Next, we observe that bragging about eating or sharing food such as \textit{coffee}, \textit{lunch}, \textit{pizza} and \textit{beer} is popular at all times except at night (see example T3).  
\begin{quote}
    \footnotesize
    T3: \textit{``Just made some lamb burgers with homemade tzatziki sauce. Starting to feel confident about my cooking skills.''}.
\end{quote}
Also, bragging in the morning is usually about things that happened the previous day (e.g., \textit{yesterday}, \textit{sleep}) or study/work (e.g., \textit{school}, \textit{class}, \textit{``Secured a B in my principles of marketing class this semester!!!''}) while bragging in the afternoon or evening involves a wide variety of recreational activities (e.g., \textit{shopping}, \textit{fitness}, \textit{\#music}, \textit{``Finally, newest member of planet fitness!''}). 
Finally, many users tend to brag about their upcoming activities in the evening (e.g., \textit{tomorrow}, \textit{morning}).

\subsection{Bragging Prevalence and Demographic Factors}

Next, we explore the relationship between bragging rate and user demographic traits. To understand what types of users brag more than others in U.S. Twitter, we perform a correlation analysis by computing the Pearson correlation coefficient between user traits and the user-level bragging metric described in Equation~\ref{equ:individualUsers} and Equation~\ref{equ:individuaAverage}. The results are summarized in the top part of Table~\ref{t:correlations}.

\renewcommand{\arraystretch}{1.2}
\begin{table}[!t]
\footnotesize
\center
\begin{tabular}{|l|c|c|c|}
\hline
\bf Trait & \bf r & \bf $p_{unc}$ & \bf $p_{corr}$\\ \hline
\multicolumn{4}{|l|}{User Demographics} \\
\hline
Gender (\female-1, \male-0) & 0.10 & $<$.001 & $<$.001 \\
Age & -0.16 & $<$.001 & $<$.001\\
Education & 0.14 & $<$.001 & $<$.001\\
Income & 0.07 & $<$.003 & $<$.002\\
\hline
\multicolumn{4}{|l|}{User Popularity} \\
\hline
No. Followers & 0.12 & $<$.001 & $<$.001 \\
No. Friends/Followers & -0.10 & $<$.001 & $<$.001 \\
No. Listings & 0.09 & $<$.001 & $<$.001\\
\hline
\end{tabular}
\caption{Pearson correlations between user-level traits and their bragging metric. $p_{unc}$ and $p_{corr}$ refer to uncorrected and corrected (Bonferroni correction) p-values.}
\label{t:correlations}
\end{table}

We first analyze gender and age. According to our analysis, gender and age are strongly associated with the bragging percentage ($p < .001$). Note that correlations around 0.1 with such a large sample size (N=2,685) are highly significant and in terms of magnitude in line with correlations between other well known linguistic variables and traits~\cite{carey2015narcissism,holgate2018swear}.

Thus, we examine the rest of the demographic traits by controlling for gender and age using partial correlation~\cite{preotiuc-pietro-etal-2015-role}, where education level and annual income are represented in the ordinal scale described in Section~\ref{ssec:demographic}. The main findings are:
\begin{itemize}
    \item Gender is significantly correlated with bragging in the sense that female users brag more than male users ($r = 0.10$, $p < .001$). This is consistent with the findings of previous studies related to self-presentation~\cite{sheldon2013examining,wang-etal-2021-self} and it can be explained by the fact that women show more interest in developing friendships online~\cite{holmes2013women}, which can be accomplished by positive self-presentation.
    
    \item There is a significant association between age and bragging, with younger users bragging more than older ones ($r = -0.16$, $p < .001$). This might result from younger people's desire to increase their status among peers and peer approval~\cite{macisaac2018she}. Social comparison was found to occur more frequently in younger age groups than in older ones~\cite{mcandrew2012does}, which explains why younger users tend to create positive self-presentations online~\cite{yau2019s}.   
    \item Users with higher education levels tend to brag more ($r = 0.14$, $p < .001$). This is might be explained by the fact that users with a higher educational level tend to express more joy which could include self-disclosure statements~\cite{volkova2015predicting}.
    
    \item Users with higher income significantly brag more frequently online ($r = 0.07$, $p < .003$) than users with lower incomes. Users with higher income were found to be more likely to produce positive tweets~\cite{volkova2015predicting}. Previous work also suggested that rich people are characterized by a self-focused and narcissistic personality~\cite{leckelt2019rich}. This causes them to produce more content that is related to self-promotion in social media~\cite{buffardi2008narcissism, moon2016role}. 
    
    \item Higher income and education are positively correlated with older age (income: $r = 0.16$, $p < .001$; education: $r = 0.30$, $p < .001$). Older age however has an inverse relationship to bragging than income and education. This highlights a divergence along these traits, where users who are either highly educated or young are likely to brag more.
\end{itemize}

We further explore the relationship between language use and bragging behavior across different demographic characteristics.

\paragraph*{Gender} Table~\ref{t:feature_gender_age} (left column) shows the top-25 unigrams correlated with gender. 
We observe that male users mostly brag about their partners (e.g., \textit{wife}), but also mention other users (e.g., \textit{$<$user$>$}). As illustrated in example T4, popular bragging topics for males are entertainment achievements such as games (e.g., \textit{\#dnd}, \textit{\#league}, \textit{\#twitch}, \textit{playing}) and sports (e.g., \textit{tournament}, \textit{win}, \textit{football}, \textit{championship}, \textit{teams}). 
\begin{quote}
    \footnotesize
    T4: \textit{``I'm so genuinely happy I witnessed that. \#NationalChampionship''}.
\end{quote}
On the other hand, female users prefer to brag with first person pronouns (e.g., \textit{my}, \textit{me}, \textit{i}). They also brag about personal traits (e.g., \textit{hair}), feelings (e.g., \textit{love}, \textit{happy}) and their partners (e.g., \textit{boyfriend}). For example, a female user mentions her haircut appointment in example T5. 
\begin{quote}
    \footnotesize
    T5: \textit{``HOORAY!! I finally got a haircut appointment!! It'll be 64 days since my last cut (I normally go every 5 weeks so this'll be close to double my normal tim, and yeah, it feels like I have about twice as much hair to cut!!)''}.
\end{quote}
According to \citet{mcandrew2012does}, women spend more energy than men in presenting themselves for impression management. Furthermore, bragging by females usually contains female-related terms (e.g., \female) or positive emoticons (e.g., :), :p) and emojis (e.g., \smilingFaceWithHeartEyes{}{}, \redHeart{}{}) to strengthen the meaning of their posts. This corroborates findings that positive emojis are used more frequently in positive contexts~\cite{derks2008emoticons}. Results are also consistent with the observation that female users communicate using more emotional exchanges~\cite{gefen2005if}. They also use emojis more often than men~\cite{chen2017through,prada2018motives}.

\renewcommand{\arraystretch}{1.3}
\begin{table}[!t]
\center
\resizebox{0.48\textwidth}{!}{
\begin{tabular}{|l|c|l|c||l|c|l|c|}
\hline
\multicolumn{4}{|c|}{\bf Male vs. Female} & \multicolumn{4}{|c|}{\bf Born after 1988 vs. before 1988} \\
\hline
Unigram & r & Unigram & r & Unigram & r & Unigram & r \\ 
\hline
$<$url$>$ & .073 & my & .083 & i & .099 & $<$url$>$ & .141\\
$<$user$>$ & .071 & so & .063 & my & .069 & $<$user$>$ & .079\\
game & .061 & \smilingFaceWithHeartEyes{}{} & .060 & \smilingFaceWithHeartEyes{}{} & .069 & ! & .077\\
team & .047 & :) & .059 & me & .057 & join & .036 \\
games & .040 & :p & .058 & so & .050 & \#livepd & .033\\
\#dnd & .036 & \redHeart{}{} & .048 & \loudylyCryingFace{}{} & .050 & local & .032\\
\#league & .036 & hair & .048 & m & .049 & kids & .031\\
wife & .034 & mom & .043 & \faceWithTearsOfJoy{}{} & .047 & our & .031\\
tournament & .034 & \smilingFace{}{} & .042 & \smilingFaceWithSmilingEyes{}{} & .041 & we & .031\\
podcast & .033 & \upsideDwonFace{}{} & .041 & class & .040 & wife & .031\\
win & .033 & bed & .041 & exam & .039 & daughter & .030\\
\#livepd & .033 & love & .040 & college & .038 & via & .029\\
football & .033 & \faceWithTearsOfJoy{}{} & .040 & semester & .038 & w & .029\\
\#twitch & .033 & \smilingFaceWithSmilingEyes{}{} & .040 & lol & .037 & play & .028\\
fantasy & .033 & me & .039 & myself & .036 & \#dnd & .028\\
great & .032 & \flushedFace{}{} & .037 & life & .036 & pe & .028\\
stream & .032 & \female & .036 & \smilingFace{}{} & .036 & inboxdollars & .028\\
aw\_prints & .031 & \personTippingHand{}{} & .035 & ve & .035 & app & .028\\
inboxdollars & .031 & \beamingFaceWithSmilingEyes{}{} & .034 & best & .035 & challenge & .027\\
championship & .030 & \loudylyCryingFace{}{} & .034 & \upsideDwonFace{}{} & .035 & covet & .026\\
playing & .030 & \seeNoEvilMonkey{}{} & .033 & \pleedingFace{}{} & .035 & show & .026\\
teams & .030 & because & .033 & \personTippingHand{}{} & .035 & awesome & .025\\
congratulations & .029 & i & .033 & \grinningFaceWithSweat{}{} & .035 & \#positive & .025\\
app & .029 & happy & .033 & like & .034 & photo & .025\\
bout & .029 & boyfriend & .033 & mom & .034 & \#i\_am & .024\\
\hline
\end{tabular}}
\caption{Unigram feature correlations with bragging between gender and age, sorted by Pearson correlation (r). All correlations are significant at $p < .001$, two-tailed t-test.}
\label{t:feature_gender_age}
\end{table}

\paragraph*{Age} In terms of age, we split the users into two age groups (born before 1988 and after 1988) using the population median. Table~\ref{t:feature_gender_age} (right column) shows the top-15 unigram features correlated with age. 
We notice that younger users (born after 1988) brag more about themselves (e.g., \textit{i}, \textit{my}, \textit{me}, \textit{myself}) and school life (e.g., \textit{class}, \textit{exam}, \textit{college}, \textit{semester}). Example (T12) indicates how a younger user brags about his/her exam result.
\begin{quote}
    \footnotesize
    T12: \emph{``Also somehow got a 90 on my soils exam... how...''}.
\end{quote}
Also, they use more emojis in their bragging posts, which is consistent with the fact that in general younger users tend to use more emojis than older ones do in social media~\cite{prada2018motives}.

The group of users above median age brag more about collective activities (e.g., \textit{our}, \textit{we}) and their affiliation such as family members (e.g., \textit{kids}, \textit{wife}, \textit{daughter}), which suggests older people are more family-focused and engage more with family activities~\cite{mcandrew2012does}. In example (T13), a user in this group mentions his/her daughter while bragging.
\begin{quote}
    \footnotesize
    T13: \emph{``My daughter is hands down the coolest person I know!''}.
\end{quote}

\paragraph*{Education \& Income:} Table~\ref{t:feature_education_income} (left column) presents the top 15 unigrams in bragging posts correlated with higher and lower education. 
We observe that people with higher levels of education use a smaller number of emojis. As illustrated in example (T14), they brag more about their jobs (e.g., \textit{student}, \textit{conference}) and activities involving food (e.g., \textit{beer}, \textit{delicious}). 
\begin{quote}
    \footnotesize
    T14: \textit{``Glad I've been working on my Adobe Suite skills this year. I'm going to be making a very special obituary poster for my uncle's celebration of life to highlight testimonials from friends and family \purpleHeart{}''}.
\end{quote}
Furthermore, people with higher education mention others (e.g., \textit{<user>}, \textit{our}, \textit{their}) when bragging. As illustrated in example (T15), This is in contrast with lower educated users who focus on their personal traits, possessions or activities (e.g., \textit{i}, \textit{my}, \textit{look}, \textit{got}). 
\begin{quote}
    \footnotesize
    T15: \textit{``I look cute as hell in this hoodie''}.
\end{quote}

\renewcommand{\arraystretch}{1.2}
\begin{table}[!t]
\center
\resizebox{0.48\textwidth}{!}{
\begin{tabular}{|l|c|l|c||l|c|l|c|}
\hline
\multicolumn{4}{|c|}{\bf Higher vs. Lower Education} & \multicolumn{4}{|c|}{\bf Higher vs. Lower Income}\\
\hline
Unigram & r & Unigram & r & Unigram & r & Unigram & r\\
\hline
$<$user$>$ & .062 & i &.074 & $<$user$>$ & .061 & :) & .043\\
students & .040 & lol & .059 & $<$url$>$ & .051 & my & .030\\
$<$url$>$ & .039 & my & .055 & aw\_prints & .032 & lol & .030\\
our & .030 & \smilingFaceWithHeartEyes{}{} & .043 & $\star$congrats & .030 & got & .026\\
$\star$congrats & .029 & m & .038 & \#iteachk & .029 & \#livepd & .023\\
pe & .025 & \okHand{}{} & .035 & pe & .029 & work & .023\\
ss & .025 & look & .033 & \#piano & .029 & cory & .022\\
season & .025 & \faceWithTearsOfJoy{}{} & .032 & $\star$\#hc & .027 & covet & .022\\
zak & .025 & got & .032 & students & .026 & renee & .021\\
conference & .024 & \beamingFaceWithSmilingEyes{}{} & .031 & \#music & .024 & makes & .021\\
beer & .024 & \smilingFace{}{} & .030 & team & .024 & boyfriend & .021\\
\#piano & .023 & $\star$\#10bw & .028 & golf & .024 & yay & .021\\
their & .023 & \smilingFaceWithSmilingEyes{}{} & .027 & win & .024 & m & .020\\
delicious & .023 & baby & .027 & \blueHeart{}{} & .023 & then & .019\\
student & .021 & \loudylyCryingFace{}{} & .026 & bet & .023 & kiss & .019\\
\hline
\end{tabular}}
\caption{Unigram feature correlations with bragging between higher and lower education level annual income, sorted by Pearson correlation (r). All correlations are significant at $p < .001$, two-tailed t-test. $\star$congrats, $\star$\#10bw, $\star$\#hc is congratulations, \#10billionwives, \#happyclassrooms.}
\label{t:feature_education_income}
\end{table}

Similar income and education levels share a similar language (e.g. \textit{$<$user$>$}, \textit{students}, \textit{congratulations} in higher education and higher income, \textit{my}, \textit{lol}, \textit{baby}, \textit{boyfriend} in lower education and lower income) (see Table~\ref{t:feature_education_income} right column for the top 15 unigrams in bragging posts correlated with higher and lower income). In addition, bragging expressions from higher income users are more related to entertainment events such as \textit{\#music}, \textit{golf} compared with higher education (see example (T16)).  
\begin{quote}

    T16: \textit{``Great golf lesson with <USER> at the PAGA! Already see the results. Now-practice to see them more often. \#GonnaBreak80''}.
\end{quote}

\subsection{Bragging Prevalence and User Popularity}

Finally, we investigate whether bragging prevalence is related to user popularity on Twitter. Table~\ref{t:correlations} shows that users who are more popular on Twitter are likely to brag more across all three popularity metrics ($r$ between $0.09$ and $0.12$, $p < .001$), when controlled for gender and age.

It is possible that users with many followers (e.g., micro- or macro-influencers) tend to interact with and try to maintain followers or obtain more followers~\cite{guo2020managing} by establishing a positive social image through bragging. It is expected that users with higher follower counts are less likely to have an audience that knows them personally and thus they aim to shape their perception through self-presentation strategies. This is in contrast with the users who have fewer friends and followers and who do not feel the need to use self-presentation strategies such as bragging to their close network. Past research showed that users are more willing to share content with positive sentiment with people that share a weak relationship (e.g., online followers) than with actual real-life friends~\cite{bak2012self}.

The linguistic feature analysis did not result in any clear patterns or meaningful insights into the expression of bragging in terms of user popularity.

\section{Conclusion}

We have presented the first large-scale quantitative study of bragging on social media, focusing on understanding the prevalence, temporal dynamics and user factors impacting bragging prevalence. Our analysis of more than 1 million English Twitter posts from users in the U.S. showed that bragging slowly becomes less prevalent with time, it is more prevalent during the weekends and daytime and that females, younger users, users with higher education, income and popularity tend to brag more. Furthermore, we uncovered using linguistic methods the prevalent topics distinctive of each user group.

For future work, we plan to compare how bragging is perceived towards building a positive self-image, reactions to bragging and cultural differences in bragging behavior across locations and languages.

\section{Ethics Statement}
Our work has received approval from the Ethics Committee of our institution and complies with the Twitter data policy for research.\footnote{https://developer.twitter.com/en/developer-terms/agreement-and-policy}

\nocite{*}
\section{Bibliographical References}\label{sec:reference}

\bibliographystyle{lrec-coling2024-natbib}
\bibliography{lrec-coling2024-example}


\end{document}